%% file: main.tex
\documentclass[letterpaper, 10 pt, conference]{ieeeconf}  

\IEEEoverridecommandlockouts                              
\usepackage{amsmath}
\usepackage{graphicx}
\usepackage[caption=false]{subfig}
\usepackage{multirow}
\usepackage{dblfloatfix}
\usepackage{siunitx}
\usepackage{booktabs}
\usepackage{longtable}

\usepackage{authblk}

\usepackage{cite}
\usepackage{nomencl}
\makenomenclature
\usepackage[table,xcdraw]{xcolor}
\usepackage[caption=false]{subfig}
\usepackage{multirow}\usepackage{graphicx}

\title{\LARGE \bf
TiltXter: CNN-based Electro-tactile Rendering of Tilt Angle for Telemanipulation of Pasteur Pipettes
}

\author{Miguel Altamirano Cabrera, 
Jonathan Tirado,
Aleksey Fedoseev,
Oleg Sautenkov,\\
Vladimir Poliakov,
Pavel Kopanev,

and Dzmitry Tsetserukou

\thanks {The authors are with the Intelligent Space Robotics Laboratory, Center for Digital Engineering, Skolkovo Institute of Science and Technology (Skoltech), 121205 Bolshoy Boulevard 30, bld. 1, Moscow, Russia. { \{m.altamirano, jonathan.tirado, aleksey.fedoseev, oleg.sautenkov, vladimir.poliakov, pavel.kopanev, d.tsetserukou\}@skoltech.ru}}
}

\begin{document}
\maketitle
\thispagestyle{empty}
\pagestyle{empty}

\begin{abstract}

The shape of deformable objects can change drastically during grasping by robotic grippers, causing an ambiguous perception of their alignment and hence resulting in errors in robot positioning and telemanipulation. Rendering clear tactile patterns is fundamental to increasing users' precision and dexterity through tactile haptic feedback during telemanipulation. Therefore, different methods have to be studied to decode the sensors' data into haptic stimuli. This work presents a telemanipulation system for plastic pipettes that consists of a Force Dimension Omega.7 haptic interface endowed with two electro-stimulation arrays and two tactile sensor arrays embedded in the 2-finger Robotiq gripper. We propose a novel approach based on convolutional neural networks (CNN) to detect the tilt of deformable objects. The CNN generates a tactile pattern based on recognized tilt data to render further electro-tactile stimuli provided to the user during the telemanipulation. The study has shown that using the CNN algorithm, tilt recognition by users increased from 23.13\% with the downsized data to 57.9\%, and the success rate during teleoperation increased from 53.12\% using the downsized data to 92.18\% using the tactile patterns generated by the CNN.

\end{abstract}

\section{Introduction}

Nowadays, teleoperation and telepresence are two of the most natural applications of haptics. The most effective combinations of various types of haptic feedback remain the subject of active research. 

The designs of tactile displays were extensively investigated in several projects. Tachi et al. \cite{Tachi2015}, \cite{Tachi2} introduced the concept of a highly immersive and mobile telexistence system, TELESAR, which includes an autostereoscopic 3D display for visual feedback, a wearable tactile display, and a thermal display. Fani et al. \cite{Fani} developed a system where force feedback was used to improve the telemanipulation of a robotic arm in a dexterous operation, and in \cite{Linkglide-s} a system for telemanipulation with tactile patterns rendered on the palm. Yem et al. \cite{Yem2017} developed FinGAR, a wearable tactile device using mechanical and electrical stimulation for fingertip interaction with the virtual world. Furthermore, Peruzzini et al. \cite{Peruzzini2012} used electro-stimulation to deliver the sensation of roughness, slickness, and coarse texture of materials (e.g., wood, paper, fabric) to the fingertip. Tirado et al. \cite{Tirado2020} proposed a tactile sharing system, ElectroAR, for remote training of human hand skills. This system consists of a tactile sensing glove on the follower's side that records pressure data about grasped objects and an electro-tactile stimulation glove on the remote side.

\begin{figure}
 \centering
 \includegraphics[width=0.4\textwidth]{ 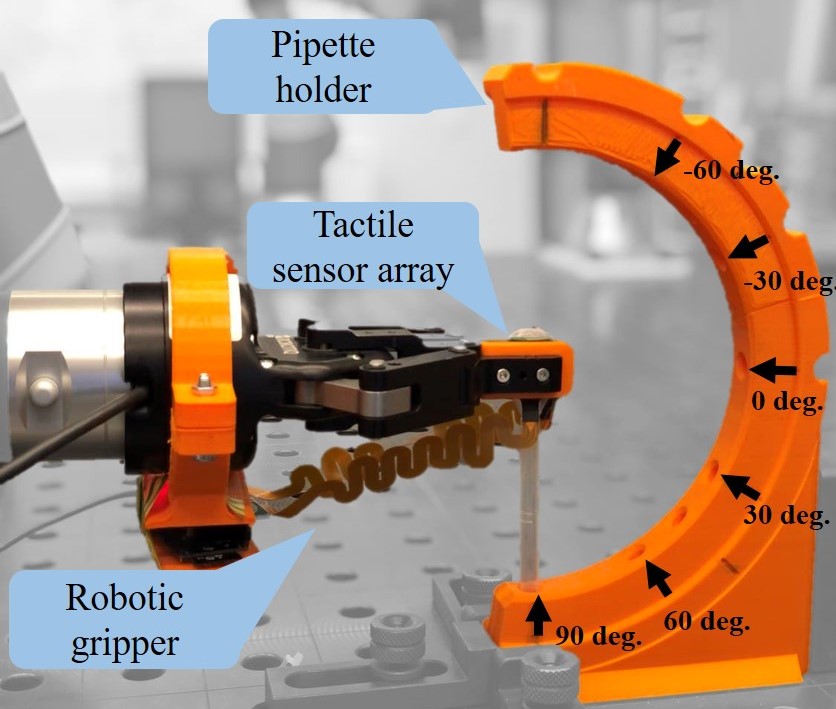}
 \vspace{-0,25 cm}
 \caption{Tilt angle recognition by the TiltXter system with high-density tactile sensor array in the remote site.}
 \label{fig:setup}
 \vspace{-0.45cm}
\end{figure}

Earlier, Sarakoglou et al. \cite{Sarakoglou2012} presented a model-based tactile feedback system with a compact 4x4 pin matrix mounted on a force feedback master device. Salvietti et al. proposed the wearable haptic thimble device for cooperative teleoperation of multiple robots \cite{Salvietti2019}. Such devices are mounted on the tracking hand's thumb and index fingers, allowing robot manipulators to be tracked by each finger motion and providing shear force feedback.
Artificial intelligence can potentially bring a new quality to telemanipulation. Some applications were developed to improve robotic perception capabilities, such as object recognition 
\cite{Gandarias} and texture recognition \cite{Nipun_2022}, \cite{Shibata}. 

In this paper, we present a novel Convolutional Neural Networks (CNN)-based telemanipulation system for deformable Pasteur pipettes. The system consists of a Force Dimension Omega.7 haptic device endowed with two electro-stimulation arrays as a master device and two tactile sensor arrays embedded in a 2-finger Robotiq gripper at the remote site. The recognized tilt generates a tactile pattern from the sensed data, and it is rendered to aid the user's tilt perception for dexterous manipulation.
This work contributes to the challenge of the fulfillment of clinical tests in remote places employing robotic systems, increasing the safety of the medical staff. In this study, a Pasteur pipette for liquid transfer (total length: $150\ mm$, total capacity: $7\ ml$, material: low-density polyethylene) was chosen because of its deformation while pressure is applied by the robotic gripper.
Two experiments were carried out to assess the performance of the approach. Firstly, we analyzed the human perception of the object tilt rendered by electro-stimulation implementing the downsizing method and by the tactile patterns generated by the CNN algorithm. Secondly, we evaluated the users' performance during the telemanipulation of plastic pipettes.

\section{TiltXter System Overview}

\begin{figure}
 \centering
 \vspace{0.15cm}
 \includegraphics[width=0.35\textwidth]{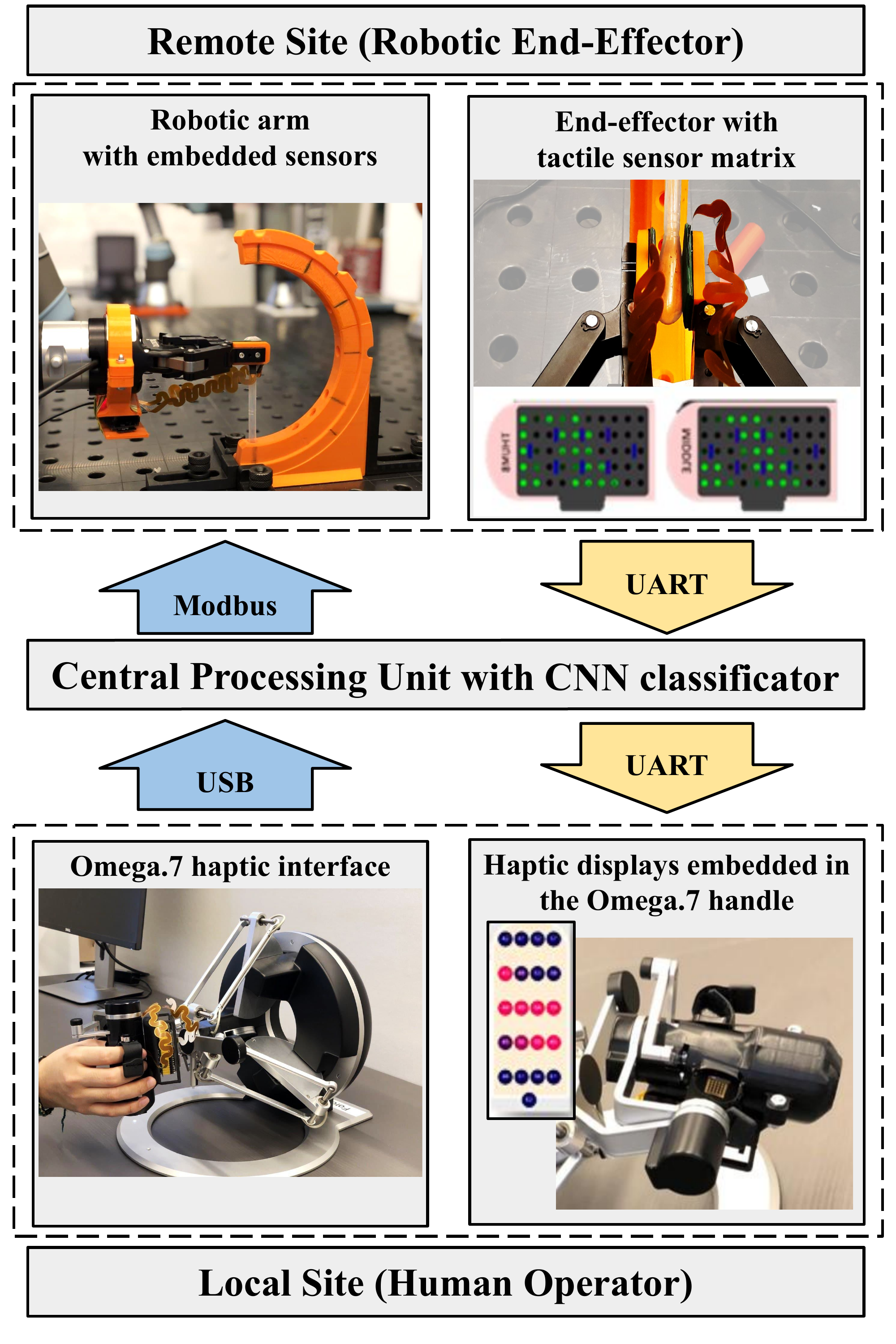}
 \caption{The overall architecture of the telemanipulation system. Green arrows define a hardware integration; blue arrows define a control signal; yellow arrows define a feedback loop.}
 \label{fig:system_architecture}
 \vspace{-0.50cm}
\end{figure}

The exterior of the system and the overall architecture are depicted in Fig. \ref{fig:setup} and Fig. \ref{fig:system_architecture}, respectively.
The user operates the Force Dimension Omega.7 haptic interface to control the 2F Robotiq gripper, mounted on the end-effector of the UR3e robotic arm. In addition, an electro-tactile display of the design proposed by Yem et al. \cite{Kajimoto-Vibol} is mounted on the gripper of the desktop haptic device to deliver the tactile perception of remote objects to the thumb and index fingers of the operator. The electro-tactile display consists of two pads with 4x5 electrode grids, which are used to render the tactile feedback from the tactile sensor arrays mounted on the tip of the gripper.
The 2-finger robotic gripper is embedded with high-density tactile sensor arrays, one on each fingertip, which allow measurement of the local pressure at each point.

The architecture of the system contains five nodes. All processes are supervised by a PC (Intel i7-7700HQ CPU, 2.8GHz, 8.15 GB of RAM, GeForce GTX 1070, running Ubuntu 18.04), which also executes the CNN node and maintains communication between the nodes. The software architecture is based on the ROS Melodic framework. The sample time for the whole process is set at $16.67\ ms$ ($60\ Hz$). During this time, the system senses the real objects, analyzes the sensed data, and delivers the electro-stimulation to the user's fingertips.

The Omega.7 controller node reads the position of the haptic display gripper and sends it to the robotic gripper.
The gripper controller node encodes the aforementioned values into commands and sends the packages to the Robotiq gripper over the Modbus interface.

\subsection{High-Fidelity Tactile Sensory and Haptic Interface}

A single-electrode array can sense an area of $5.8\ cm^2$ with a resolution of 100 points per frame. The sensor response frequency is $120\ Hz$. The range of force detection per point is $1-9\ N$. Thus, the robot detects the pressure applied to solid or flexible objects grasped by robotic fingers with a resolution of 200 points (100 points per finger). 
To deliver tangible sensations to the operator, an intuitive haptic interface that renders the shape and orientation of the object must be configured. Thus, we added a multi-contact electro-tactile display to the gripper of the Omega.7 haptic device to provide the perception of real-world object surfaces grasped by the robotic fingers. 
The electro-tactile stimulation display uses electrodes that transmit electrical current to the skin. The current transmission causes the activation of skin mechanoreceptors and, hence, the perception of tactile sensations. Two electrode arrays were located in the Omega.7 gripper, one for each of the user's fingers, with a resolution of 20 electrodes each, which covers a maximum frame area of $1.44\ cm^2$. Each electrode can provide a maximum stimulus intensity of $10\ mA$ with a base stimulation frequency of $120\ Hz$.

\section{Electro-tactile Rendering}

The first stage of the rendering method resizes the sensor data array (10x10 cells), reducing it to the electrode array size (5x4 cells). We used the bicubic interpolation algorithm to downsize tactile information. This mathematical resampling algorithm produces a smooth output array with few interpolation artifacts \cite{Ethan}, while maintaining a low computational cost. This stage results in digital tactile patterns that preserve most of their original contact surface details and permit a more natural recognition of artificial tactile patterns.

\begin{figure}[h!]
 \centering
 \includegraphics[width=0.4\textwidth]{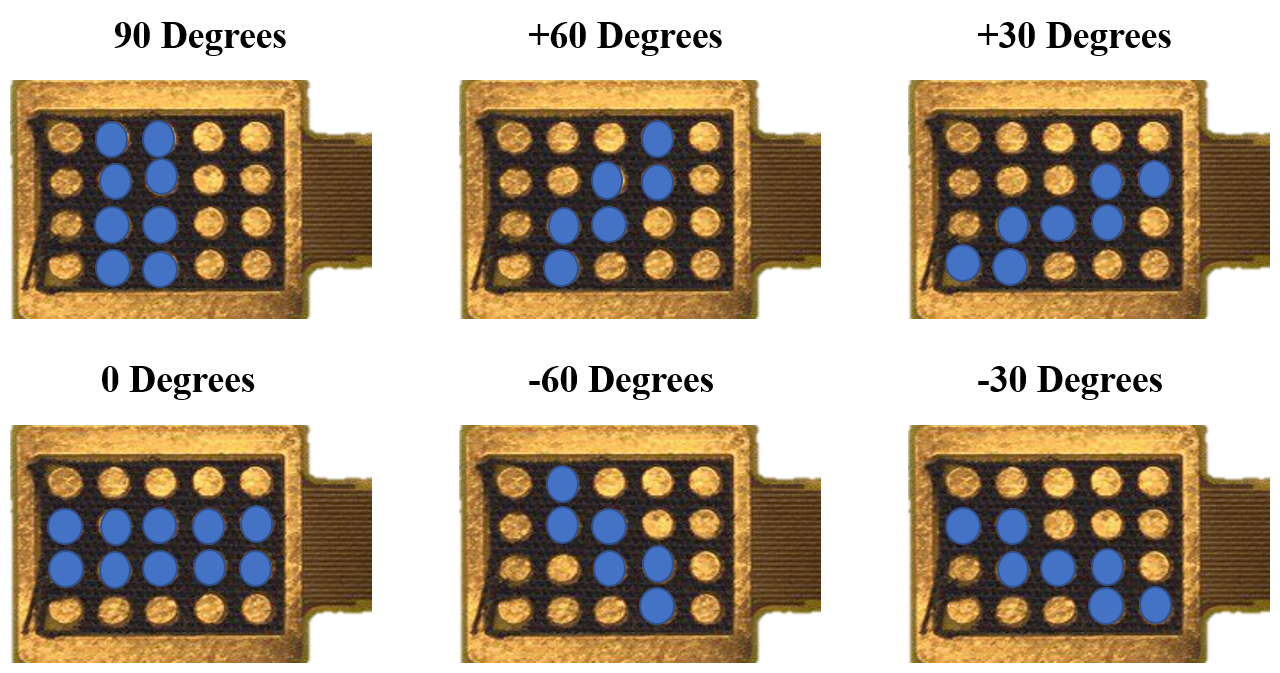}
 \caption{Set of tactile patterns represented on the electrode array corresponding to the index finger distribution.}
 \label{fig:finger_patterns}
 \vspace{-0.35cm}
\end{figure}

The second stage proposes to use a set of predefined tactile pattern arrays (Fig. \ref{fig:finger_patterns}). The use of a specific tactile pattern depends on CNN tilt estimation. The final tactile stimuli will be the Boolean multiplication (AND) between the tactile pattern array and the downsized array generated in the previous stage. The final result of this operation will be the tactile stimulus delivered by the arrays of electro-stimulation display. 

\subsection{CNN Architecture Description}

\begin{figure}[h!]
    \centering
    \vspace{-0.4cm}
    \includegraphics[width=0.47\textwidth]{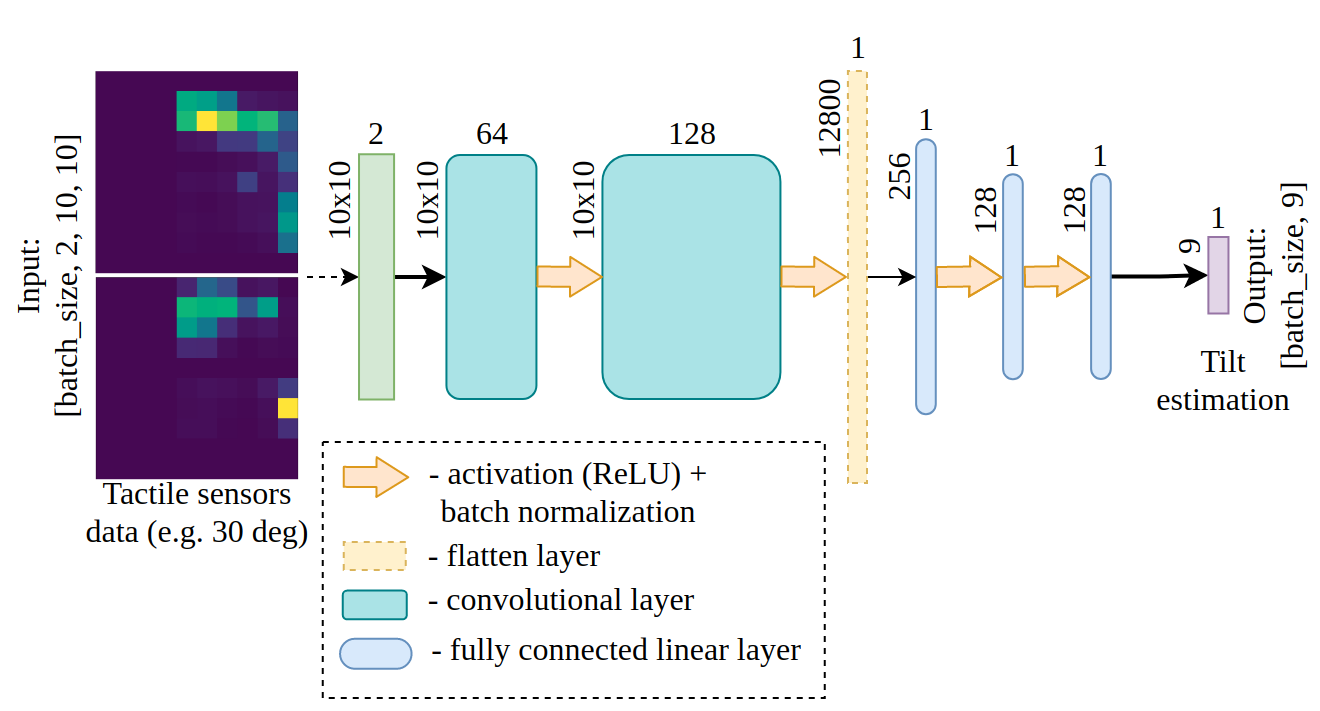}
    \caption{CNN model architecture for tilt angle classification from the tactile sensor data.}
    \label{ARCH}
    
\end{figure}

CNN has found applications in many areas, but commonly they are used for image processing. Nevertheless, it is possible to use CNNs for tactile sensor data processing. Gandarias et al. \cite{CNN_Tactile} suggested using CNN with high-resolution tactile sensors for object recognition. In \cite{Arapi}, Arapi et al. predicted the failures during grasping by utilizing the information from an array of Inertial Measurement Units. Arimatsu et al. \cite{SONY} applied CNNs in next-generation virtual reality controllers for finger tracking. Therefore, it has been proven that CNN can be used in classification and regression tasks with tactile data. In this paper, we present a convolutional neural network (CNN) model for pipette tilt recognition. The architecture of the proposed CNN model is illustrated in Fig. \ref{ARCH}.

The network consists of two convolutional layers and four fully connected linear layers with ReLU activation functions and batch normalization, which speeds up training and enables faster and more effective optimization.

\subsection{Model training and validation}

\begin{figure}[h!]
    \centering
    \vspace{0.25cm}
    \includegraphics[width=0.46\textwidth]{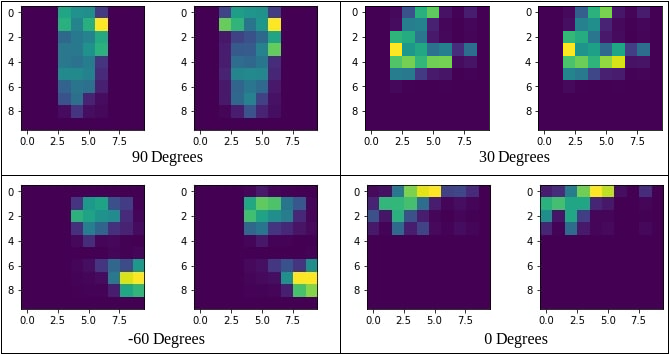}
    \caption{Data samples from the tactile sensors on the left and right fingers of the gripper during its contact with the tilted pipette. The data from the left sensor was flipped horizontally for dataset consistency.}
    \label{DATA}
    \vspace{-0.15cm}
\end{figure}

For model training and validation, 8928 data pairs from the tactile sensor arrays were collected for 9 angle classes ($\pm90$ deg., $\pm60$ deg., $\pm45$ deg., $\pm30$ deg., $0$ deg.) and for 31 gripper positions (from the minimum to maximum pressure applied to the pipette.). In order to avoid resubstitution validation or evaluation \cite{train_test} the data was split into a training set (50\% of the dataset), a validation set (25\% of the dataset), and a test set (25\% of the dataset) for performance evaluation of the network. Data samples are shown in Fig. \ref{DATA}.

\begin{figure}[h!]
    \centering
    \includegraphics[width=0.47\textwidth]{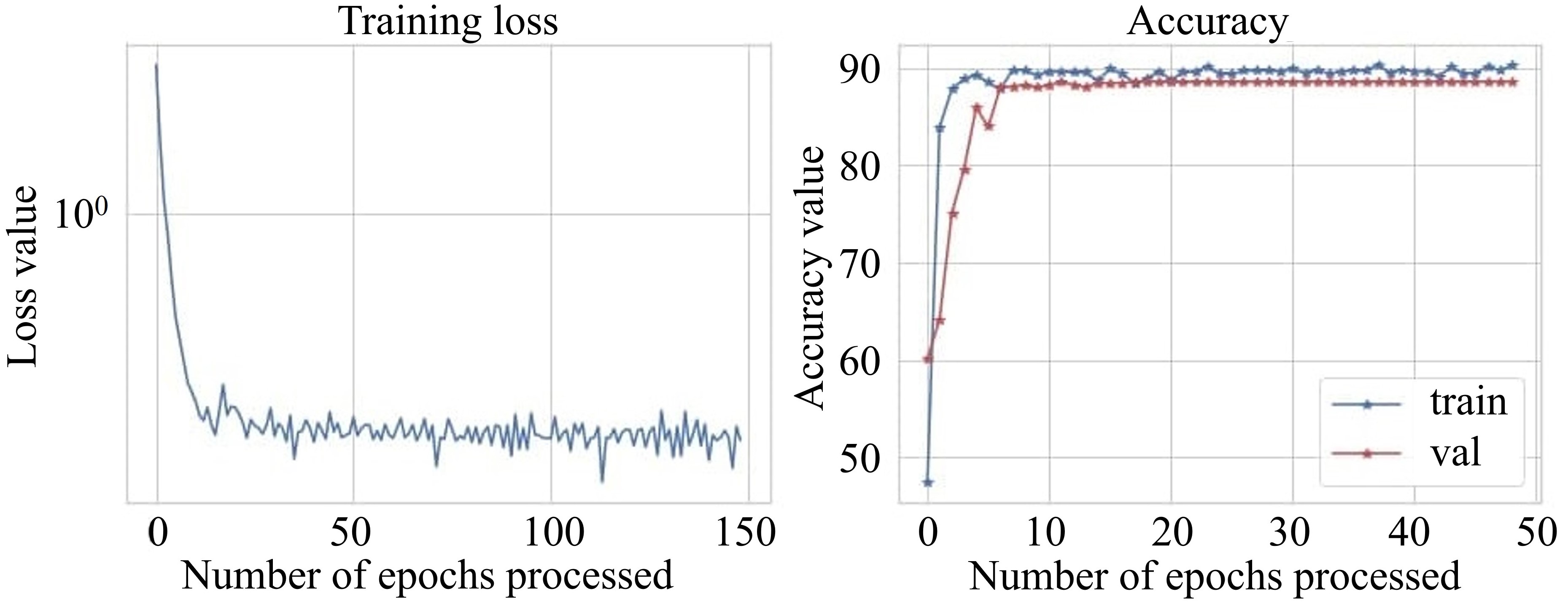}
    \caption{Learning curves of tilt angle classification with CNN.}
    \label{CURVES}
    \vspace{-0.35cm}
\end{figure}

The deep learning model was made with the PyTorch 
open-source machine learning framework and the dynamic learning rate reducing scheduler ReduceLROnPlateau.
The cross-entropy loss was applied as a criterion, as it combines the LogSoftmax function with the negative log-likelihood loss. Learning curves are shown in Fig. \ref{CURVES}. 

The tilt prediction model achieved 88.82$\%$ validation accuracy during 50 epochs of training. Model accuracy on the test dataset achieved 88.39$\%$, which is comparable with performance on the validation dataset.

\section{Experimental Evaluation of Tilt Perception}

The accuracy of object orientation perception by humans has been extensively explored in several studies. Bensmaia et al. \cite{Bensmaia2008} evaluated the human's ability to identify the tilt of bars and edges presented to the distal fingerpad with visual and tactile stimulation. Peters et al. \cite{Peters2015} investigated the means behind the human ability to resolve the orientation of edges with an experiment based on a bar set of various lengths and a tilt angle step of 60 deg. Pérez-Bellido et al. \cite{Perez-Bellido2018} proposed the hypothesis that tactile signals engage the neural mechanisms supporting visual contextual modulation, which access the neural circuits underlying the vision modulation. Such results support the importance of tactile tilt perception and suggest its potential improvement with the fixated size of the patterns generated by the electro-tactile stimulation. 

We conducted a series of user evaluation experiments to determine the extent to which electro-tactile feedback with the tactile pattern generated by the CNN can improve raw haptic feedback in telemanipulation. Fourteen participants (four females) aged 22 to 31 volunteered to take the test. None of them reported any deficiencies in sensorimotor function. The participants were informed about the experiments and agreed to the consent form. This user study was approved by the Institutional Review Board of the Skolkovo Institute of Science and Technology (MoM Protocol No. 5 dd. July 21, 2021).

\subsection{Experiment on Human Perception of Pipette Tilt Angle}

The experiment was conducted to evaluate the perception of different tilt positions described in Section III. 
The holder design allows for the location of the pipette at different angles, preventing the movement of the user's hand when the pipette's angle is modified. The pipettes were located at 6 different angles ($90$ deg., $\pm60$ deg., $\pm30$ deg., $0$ deg.).

\begin{figure}[t!]
 \centering
 \vspace{0.15cm}
 \includegraphics[width=0.35\textwidth]{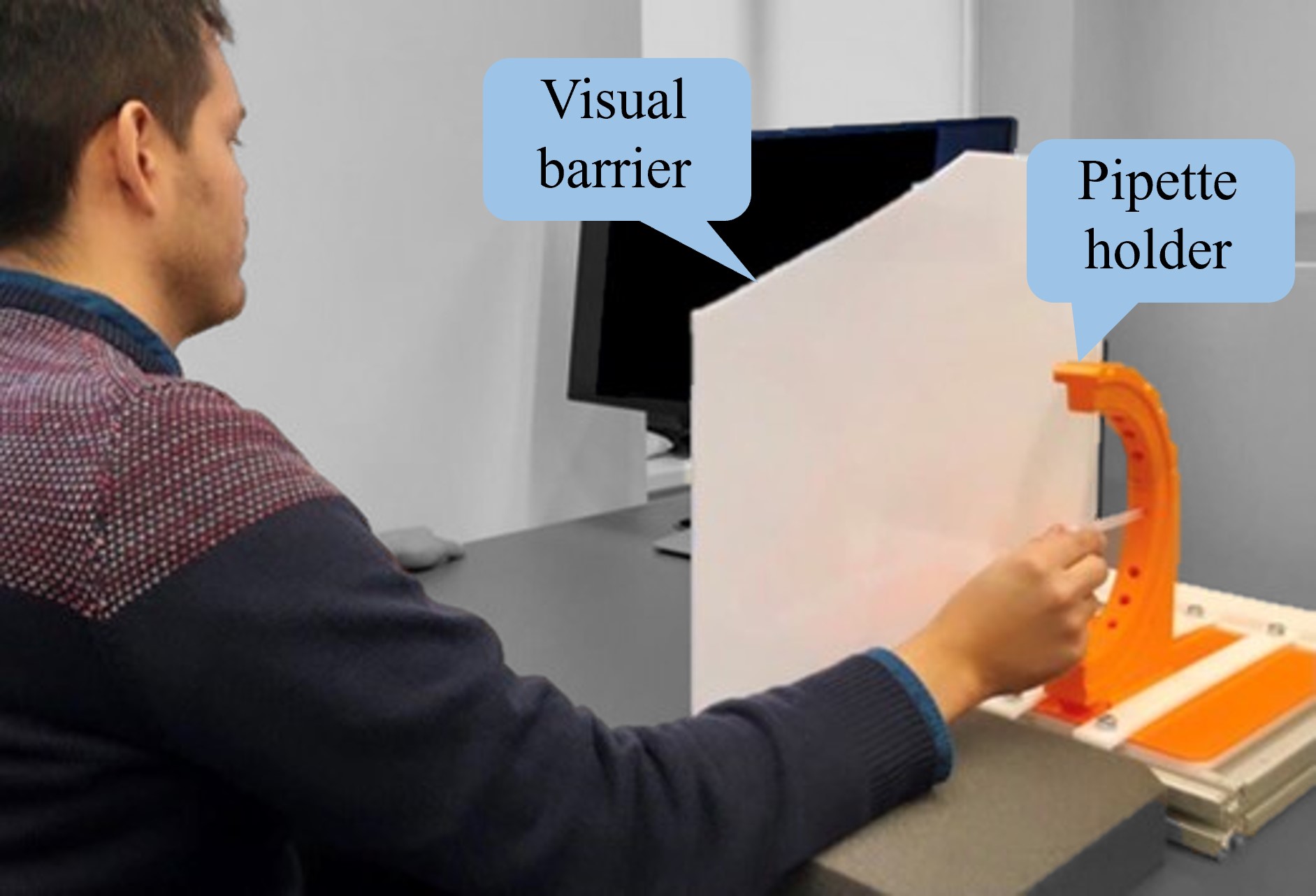}
 \caption{Experimental setup of the human perception of pipette tilt angle.}
 \label{fig:ex1_setup}
 \vspace{-0.45cm}
\end{figure}
Before the experiment, a training session was conducted where the pipettes were placed at each angle. During the experiment, the user was asked to sit in front of a desk with the occlusion barrier preventing visual contact with their right hand, as shown in Fig. \ref{fig:ex1_setup}. The pipette was placed at each tilt angle five times in random order. Each time, the user was asked to blindly evaluate the pipette angle only grasping it with his fingertips (thumb and index fingers). The screen showed a picture with the empty pipette holder presenting its corresponding angles to support the users. 
The results of the human perception of the pipette tilt angle during interaction without any proxy device are summarized in a confusion matrix (see Table \ref{CM1}).

\input{confusion_matrix_human_perception}

The perception of the angles was analyzed with a one-way ANOVA without replication with a chosen significance level of $\alpha<0.05$. The ANOVA results showed a statistically significant difference between the evaluated tilt angles of the pipette ($F(5,84) = 5.07$, $p=7.0\cdot10^{-4}$). 
The paired t-tests with one-step Bonferroni correction showed statistically significant differences between $0$ and $30$ deg. ($p=0.049$), $0$ and $60$ deg. ($p=0.04$), $0$ and $-30$ deg. ($p=0.001$), $-60$ and $-30$ deg. ($p=0.045$). The open-source statistical package Pingouin was used for the statistical analysis. 
The overall recognition rate is $77.92\%$, supporting our initial hypothesis that the human perception system is able to distinguish the angles with a tolerance of $30$ deg. The average recognition time is $5.22$ sec., with the maximum average recognition time for the $60$ degrees. ($6.64$ sec.) and the minimum average recognition time for the $0$ degree ($3.60$ sec.).

\subsection{Experiment on Electro-tactile Perception}

The second evaluation centered on the analysis of human perception of the electro-tactile rendering. During the first experiment, the tilt perception was rendered directly from sensor data to the electro-stimulation display using the downsizing method described in Section III.A. During the second experiment, we assessed participants' perception of the pipette tilt angle with the tactile pattern generated by the CNN classification described in Section III.B.
Before the study, during a training session, the experimenter explained the purpose of the multi-contact haptic interface to each participant and demonstrated the electro-tactile feedback for each of the six angles of the pipette. The demonstration was provided first with additional visual feedback and then blindly at least once. During the experiment, the user was asked to sit in front of a desk and control the Robotiq gripper by using the Omega.7 device with their right hand. 
Participants then performed a series of grasping tasks with the pipette, evaluating the angle of its tilt. Each angle was presented five times blindly in random order; thus, 30 patterns were provided to each participant in each of the evaluations.

\subsubsection{Downsized Direct Data Rendering Results}
The results of the human perception of the pipette tilt angle obtained by rendering the downsized data are summarized in a confusion matrix (see Table II).

\input{confusion_matrix_without_mask}

In order to evaluate the statistical significance of the differences between the perception of the angles with downsized direct data rendering, we analyzed the results using single-factor repeated-measures ANOVA with a chosen significance level of $\alpha<0.05$. 
According to the ANOVA results, there is no statistically significant difference in the recognition rates for the different angles: $F(5,84) = 0.5993, p = 0.7005$.
Therefore, we cannot confirm that a statistically significant difference exists between the recognized patterns. The overall recognition rate of $21.67\%$ implies that users can not distinguish the angles. The average recognition time is $9.51$ sec.

\subsubsection{Tactile Pattern Data Rendering Results}

The confusion matrix summarizes the human perception of the pipette tilt angle from the tactile pattern data rendered by CNN (see Table III).

\input{confusion_matrix_with_mask}

In order to evaluate the statistical significance of the variations in the perception of the six angles, we analyzed the results using a single-factor repeated-measures ANOVA with a chosen significance level of $\alpha<0.05$. The ANOVA results revealed a statistically significant difference in the recognition rates for the different angles, with $F(5,84) = 5.6963, p = 1.3\cdot10^{-4}$. These results indicate that the angles significantly influenced the percentage of correct responses.
The paired t-tests with one-step Bonferroni correction showed statistically significant differences between $0$ and $60$ deg. ($p=3.5\cdot10^{-3}$), $0$ and $-60$ deg. ($p=1.5\cdot10^{-3}$), $60$ and $90$ deg. ($p=1.5\cdot10^{-2}$), and $90$ and $-60$ deg. ($p=6.9\cdot10^{-3}$). The open-source statistical package Pingouin was used for the statistical analysis. 

The overall recognition rate is $57.6\%$. However, if the patterns are paired by their orientation, the overall recognition increases to $72.5\%$. This effect can be attributed to confusion between angles with the same orientation. For instance, it can be noticed in Table III, where the confusion rates between 60 and 30 deg. are $0.20\%$ and $0.24\%$, respectively, and the confusion between -60 and -30 deg. have similar rates of $0.20\%$ and $0.21\%$, respectively. The average recognition time is $6.9$ sec. 
Considering the results from the downsized direct data and the tactile patterns data rendering evaluations, a two-way ANOVA was conducted to explore the effect of the application of the tactile patterns generated by the CNN classification and the different pattern angles on the human recognition rate, which indicated that there was a significant difference between the application of the tactile patterns and the recognition rate of the angles ($p=1.98\cdot10^{-2}$). 

\subsection{Experiment on Teleoperated Grasping}

The orientation of the robotic TCP was controlled by users through the position of the Omega.7 device endowed with the electro-tactile interface at the local site (Fig. \ref{fig:setup}). The pipette was located in the 90-degree holder at the remote site. The user's task was to modify the orientation of the robot TCP and grasp the pipette two times at each angle in three different conditions: with no haptic feedback, with electro-tactile stimulation using the downsizing method, and with the tactile patterns by CNN. 

Nine participants (three females) aged 22 to 31 volunteered to conduct the test. A training session was conducted before the experiment, where the teleoperation of the robotic TCP and gripper was explained, and the feedback modalities were presented. The real-time video of the remote site was streamed on a screen. While no haptic feedback was provided, the users were asked to locate the TCP robot in the correct orientation and to close the gripper when they were ready. After that, the experimenter mentioned if the angle was grasped in the proper orientation. While the electro-tactile stimulation was active, the users were allowed to modify the angle according to the information they perceived. At the end of the experiment, the users were asked to fill out an NASA Task Load Index questionnaire (NASA-TLX) \cite{hart1987background}, to analyze each teleoperation experience as a subjective dependent variable. NASA-TLX is a workload rating procedure developed by the Human Performance Group at NASA. It measures the perceived workload of a task over six sub-scales (ranging from 1–20): mental demand, physical demand, temporal demand, subject performance, effort, and frustration. 

\subsubsection{Experimental Result}
\begin{figure}[h!]
\centering

 \includegraphics[width=0.33\textwidth]{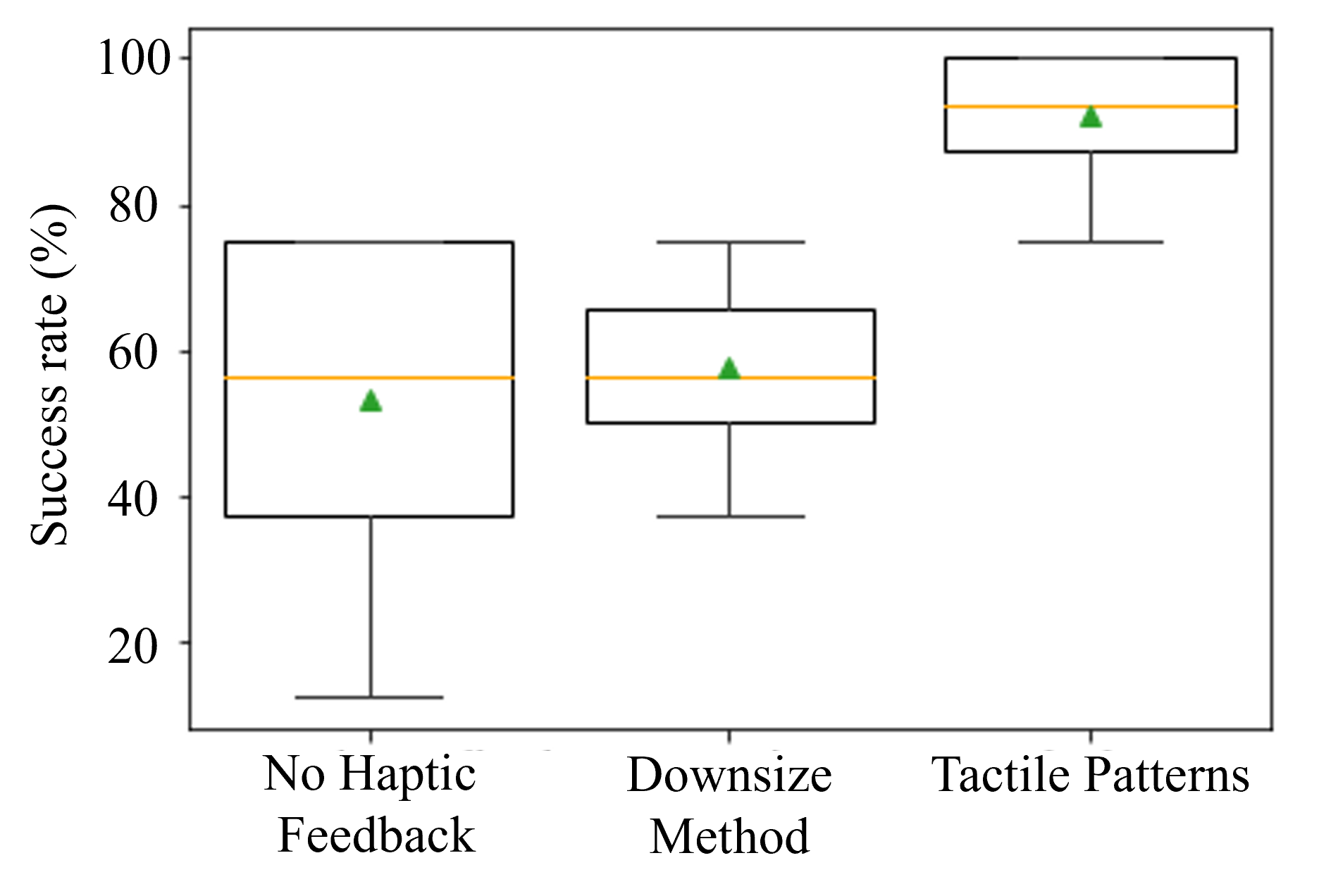}
 
  \vspace{-0.35cm}
 \caption{Success rate of angle grasping during teleoperation at different gripper orientations without haptic feedback, using the downsizing method, and the tactile patterns by CNN.}
 \label{fig:exp3_percentage}
 \vspace{-0.35cm}
\end{figure}

The success rate of angle grasping during the experiment is summarized in Fig. \ref{fig:exp3_percentage}. 

The average success rate using only visual feedback is 53.12\%, with the downsizing method 57.81\%, and using the tactile pattern generated by CNN 92.18\%. In order to evaluate the statistical significance of the differences between the implementation of the haptic feedback with the two different methods and without haptic feedback, we analyzed the results using a single-factor repeated-measures ANOVA, with a chosen significance level of $\alpha<0.05$. According to the ANOVA results, there is a statistically significant difference in the success rates for the different conditions, $F(2,21) = 13.8759, p = 14\cdot10^{-4}$. The ANOVA showed that the use of different methods for visual and haptic feedback influences the success rate of grasping during teleoperation. The paired t-test with Bonferroni correction showed a statistical difference between the use of only visual feedback and the tactile patterns method ($p=3.2\cdot10^{-5}$) and between the use of the downsizing method and the tactile patterns ($p=5.3\cdot10^{-4}$). 

The results from the NASA-TLX questionnaire are summarized in Table \ref{CM4}. Friedman tests were used to analyze the results. Statistically significant differences were found in mental demand ($p=0.0214 < 0.05$), temporal demand ($p=0.0077$), own performance ($p=0.0177$), effort ($p=0.0316$), and frustration ($p=0.0038$). 

\input{NasaTLX}

Wilcoxon tests were performed for pairwise comparisons for the five sub-scales that presented statistical differences. It was found that using both the downsizing method ($p=0.038$) and the tactile patterns method ($p=0.027$) is less mentally demanding than not using haptic feedback. In comparison to not using haptic feedback, the implementation of the tactile patterns generated by CNN decreases the temporal demand ($p=0.027$), the effort ($p=0.042$), and the frustration ($p=0.018$), and increases the own performance ($p=0.020$). In relation to the implementation of the downsizing method, the tactile pattern method decreases the temporal demand ($p=0.027$), the effort ($p=0.027$), and the frustration ($p=0.027$), and increases the own performance ($p=0.019$).

\section{Conclusions and Future Work}

In the presented work, the TiltXter tilt recognition system with CNN-based electro-tactile rendering for telemanipulation of Pasteur pipettes was introduced, and tilt tactile perception was studied. The experiments have shown that electro-tactile feedback with downsized data has significantly poorer performance than natural human perception of the object orientation. By applying the tactile patterns generated by CNN, we were able to increase the operator's perception of the tilt from $33.43\%$ to $72.5\%$, bringing it closer to natural precision. Based on this evidence, we can conclude that the use of electro-tactile feedback in combination with CNN-based rendering methods can potentially improve the telemanipulation of deformable objects.

Additionally, the teleoperated grasping experiment showed that the success rate of grasping pipettes placed at different angles increases from 53.12\% with only visual feedback to 92.18\% using tactile patterns. The NASA-TLX questionnaire revealed lower temporal demand, effort, and frustration while tactile patterns were applied in comparison with the downsized method and without haptic feedback. 

The future work of this project is to acquire an extended dataset that includes centered and non-centered object patterns scanned by the sensor arrays attached to the gripper's fingers, allowing for more accurate detection of the object's tilt at higher angular resolution. The proposed system and CNN-based rendering method can be applied to increase tilt recognition of laboratory instruments at remote co-working labs, improving dexterous telemanipulation and the users' reactions.

\section*{Acknowledgements} 
Research reported in this publication was financially supported by the RSF grant No. 24-41-02039.


\end{document}

%% file: confusion_matrix_human_perception.tex
\begin{table}[h!]
\centering{
 \vspace{-0.35cm}
\caption{Confusion Matrix for Actual and Perceived Tilt Angles Across All Subjects by the Human Perception Study.}
\label{CM1}
\setlength{\tabcolsep}{8pt} 
\renewcommand{\arraystretch}{0.95} 
\begin{tabular}{|cc|cccccc|}
\hline
\multicolumn{2}{|c|}{} &
  \multicolumn{6}{c|}{\textit{Answers (Predicted Class)}} \\ \cline{3-8} 
\multicolumn{2}{|c|}{\multirow{-2}{*}{\%}} &
 \multicolumn{1}{c|}{0} &
 \multicolumn{1}{c|}{30} &
 \multicolumn{1}{c|}{60} &
 \multicolumn{1}{c|}{90} &
 \multicolumn{1}{c|}{-60} &
 -30 \\ \hline
 \multicolumn{1}{|c|}{}&
  0 &
  \multicolumn{1}{c|}{\cellcolor[HTML]{1F4E78}{\color[HTML]{F9FAFB} 0.91}} &
  \multicolumn{1}{c|}{\cellcolor[HTML]{F3F6F8}0.05} &
  \multicolumn{1}{c|}{\cellcolor[HTML]{F9FAFB}0.00} &
  \multicolumn{1}{c|}{\cellcolor[HTML]{F9FAFB}0.00} &
  \multicolumn{1}{c|}{\cellcolor[HTML]{F9FAFB}0.00} &
  \cellcolor[HTML]{F9FBFC}0.04 \\ \cline{2-8} 
\multicolumn{1}{|c|}{}&
 
  30 &
  \multicolumn{1}{c|}{\cellcolor[HTML]{F3F6F8}0.06} &
  \multicolumn{1}{c|}{\cellcolor[HTML]{476E90}{\color[HTML]{F9FAFB} 0.75}} &
  \multicolumn{1}{c|}{\cellcolor[HTML]{DBE3EA}0.14} &
  \multicolumn{1}{c|}{\cellcolor[HTML]{F9FAFB}0.00} &
  \multicolumn{1}{c|}{\cellcolor[HTML]{F6F8FA}0.04} &
  \cellcolor[HTML]{F9FAFB}0.01 \\ \cline{2-8} 
  
 \multicolumn{1}{|c|}{}&
 
  60 &
  \multicolumn{1}{c|}{\cellcolor[HTML]{FCFDFE}0.01} &
  \multicolumn{1}{c|}{\cellcolor[HTML]{CFD9E2}0.22} &
  \multicolumn{1}{c|}{\cellcolor[HTML]{507596}{\color[HTML]{F9FAFB} 0.71}} &
  \multicolumn{1}{c|}{\cellcolor[HTML]{F3F6F8}0.05} &
  \multicolumn{1}{c|}{\cellcolor[HTML]{F9FAFB}0.00} &
  \cellcolor[HTML]{FCFDFE}0.01 \\ \cline{2-8} 
\multicolumn{1}{|c|}{}&
  90 &
  \multicolumn{1}{c|}{\cellcolor[HTML]{FCFDFE}0.01} &
  \multicolumn{1}{c|}{\cellcolor[HTML]{FCFDFE}0.01} &
  \multicolumn{1}{c|}{\cellcolor[HTML]{EAEFF3}0.08} &
  \multicolumn{1}{c|}{\cellcolor[HTML]{325D83}{\color[HTML]{F9FAFB} 0.86}} &
  \multicolumn{1}{c|}{\cellcolor[HTML]{F6F8FA}0.04} &
  \cellcolor[HTML]{F9FAFB}0.00 \\ \cline{2-8} 
\multicolumn{1}{|c|}{}&
  -60 &
  \multicolumn{1}{c|}{\cellcolor[HTML]{F9FBFC}0.03} &
  \multicolumn{1}{c|}{\cellcolor[HTML]{F9FAFB}0.00} &
  \multicolumn{1}{c|}{\cellcolor[HTML]{F9FBFC}0.01} &
  \multicolumn{1}{c|}{\cellcolor[HTML]{F6F8FA}0.04} &
  \multicolumn{1}{c|}{\cellcolor[HTML]{4A7092}{\color[HTML]{F9FAFB} 0.76}} &
  \cellcolor[HTML]{D8E0E8}0.16 \\ \cline{2-8} 
\multicolumn{1}{|c|}{\multirow{-6}{*}{\textit{\rotatebox{90}{Patterns}}}} &
  -30 &
  \multicolumn{1}{c|}{\cellcolor[HTML]{E1E8ED}0.13} &
  \multicolumn{1}{c|}{\cellcolor[HTML]{FCFDFE}0.02} &
  \multicolumn{1}{c|}{\cellcolor[HTML]{F6F8FA}0.02} &
  \multicolumn{1}{c|}{\cellcolor[HTML]{F9FAFB}0.00} &
  \multicolumn{1}{c|}{\cellcolor[HTML]{DBE3EA}0.15} &
  \cellcolor[HTML]{5C7E9D}{\color[HTML]{F9FAFB} 0.68} \\ \hline

\end{tabular}}
\vspace{-0.25cm}
\end{table}

%% file: confusion_matrix_without_mask.tex
\begin{table}[h!]
\label{table:confusion_without_mask} 
\centering{
\caption{Confusion Matrix for Actual and Perceived Tilt Angles for Recognition with Downsized Direct Data.}
\label{CM2}
\setlength{\tabcolsep}{8pt} 
\renewcommand{\arraystretch}{0.95} 
\begin{tabular}{|cc|cccccc|}

\hline
\multicolumn{2}{|c|}{} &
  \multicolumn{6}{c|}{\textit{Answers (Predicted Class)}} \\ \cline{3-8} 
\multicolumn{2}{|c|}{\multirow{-2}{*}{\%}} &
  \multicolumn{1}{c|}{0} &
  \multicolumn{1}{c|}{30} &
  \multicolumn{1}{c|}{60} &
  \multicolumn{1}{c|}{90} &
  \multicolumn{1}{c|}{-60} &
  -30 \\ \hline
\multicolumn{1}{|c|}{} &
  0 &
  \multicolumn{1}{c|}{\cellcolor[HTML]{2D5A81}{\color[HTML]{F9FAFB} 0.27}} &
  \multicolumn{1}{c|}{\cellcolor[HTML]{2D5A81}{\color[HTML]{F9FAFB} 0.27}} &
  \multicolumn{1}{c|}{\cellcolor[HTML]{F9FAFB}0.06} &
  \multicolumn{1}{c|}{\cellcolor[HTML]{7391AB}0.20} &
  \multicolumn{1}{c|}{\cellcolor[HTML]{D5DEE6}0.10} &
  \cellcolor[HTML]{D5DEE6}0.10 \\ \cline{2-8} 
\multicolumn{1}{|c|}{} &
  30 &
  \multicolumn{1}{c|}{\cellcolor[HTML]{8FA7BC}0.17} &
  \multicolumn{1}{c|}{\cellcolor[HTML]{9DB2C4}0.16} &
  \multicolumn{1}{c|}{\cellcolor[HTML]{497092}{\color[HTML]{F9FAFB} 0.24}} &
  \multicolumn{1}{c|}{\cellcolor[HTML]{6586A3}0.21} &
  \multicolumn{1}{c|}{\cellcolor[HTML]{F2F4F7}0.07} &
  \cellcolor[HTML]{ABBDCD}0.14 \\ \cline{2-8} 
\multicolumn{1}{|c|}{} &
  60 &
  \multicolumn{1}{c|}{\cellcolor[HTML]{577B9A}{\color[HTML]{F9FAFB} 0.23}} &
  \multicolumn{1}{c|}{\cellcolor[HTML]{F2F4F7}0.07} &
  \multicolumn{1}{c|}{\cellcolor[HTML]{577B9A}{\color[HTML]{F9FAFB} 0.23}} &
  \multicolumn{1}{c|}{\cellcolor[HTML]{1F4E78}{\color[HTML]{F9FAFB} 0.29}} &
  \multicolumn{1}{c|}{\cellcolor[HTML]{F2F4F7}0.07} &
  \cellcolor[HTML]{C7D3DE}0.11 \\ \cline{2-8} 
\multicolumn{1}{|c|}{} &
  90 &
  \multicolumn{1}{c|}{\cellcolor[HTML]{7391AB}0.20} &
  \multicolumn{1}{c|}{\cellcolor[HTML]{819CB4}0.19} &
  \multicolumn{1}{c|}{\cellcolor[HTML]{819CB4}0.19} &
  \multicolumn{1}{c|}{\cellcolor[HTML]{577B9A}{\color[HTML]{F9FAFB} 0.23}} &
  \multicolumn{1}{c|}{\cellcolor[HTML]{D5DEE6}0.10} &
  \cellcolor[HTML]{D5DEE6}0.10 \\ \cline{2-8} 
\multicolumn{1}{|c|}{} &
  -60 &
  \multicolumn{1}{c|}{\cellcolor[HTML]{819CB4}0.19} &
  \multicolumn{1}{c|}{\cellcolor[HTML]{C7D3DE}0.11} &
  \multicolumn{1}{c|}{\cellcolor[HTML]{E3E9EF}0.09} &
  \multicolumn{1}{c|}{\cellcolor[HTML]{8FA7BC}0.17} &
  \multicolumn{1}{c|}{\cellcolor[HTML]{497092}{\color[HTML]{F9FAFB} 0.24}} &
  \cellcolor[HTML]{7391AB}0.20 \\ \cline{2-8} 
\multicolumn{1}{|c|}{\multirow{-6}{*}{\textit{\rotatebox{90}{Patterns}}}} &
  -30 &
  \multicolumn{1}{c|}{\cellcolor[HTML]{497092}{\color[HTML]{F9FAFB} 0.24}} &
  \multicolumn{1}{c|}{\cellcolor[HTML]{ABBDCD}0.14} &
  \multicolumn{1}{c|}{\cellcolor[HTML]{B9C8D5}0.13} &
  \multicolumn{1}{c|}{\cellcolor[HTML]{7391AB}0.20} &
  \multicolumn{1}{c|}{\cellcolor[HTML]{C7D3DE}0.11} &
  \cellcolor[HTML]{8FA7BC}0.17 \\ \hline
\end{tabular}}
\vspace{-0.25cm}
\end{table}

%% file: confusion_matrix_with_mask.tex
\begin{table}[h!]
\label{table:confusion_with_mask}
\centering{
\caption{Confusion Matrix for Actual and Perceived Tilt Angles Across Subjects for Recognition of the Tactile Patterns.}
\label{CM3}
\setlength{\tabcolsep}{8pt} 
\renewcommand{\arraystretch}{0.95} 
\begin{tabular}{|cc|cccccc|}
\hline
\multicolumn{2}{|c|}{} &
  \multicolumn{6}{c|}{\textit{Answeres   (Predicted Class)}} \\ \cline{3-8} 
\multicolumn{2}{|c|}{\multirow{-2}{*}{\%}} &
  \multicolumn{1}{c|}{0} &
  \multicolumn{1}{c|}{30} &
  \multicolumn{1}{c|}{60} &
  \multicolumn{1}{c|}{90} &
  \multicolumn{1}{c|}{-60} &
  -30 \\ \hline
\multicolumn{1}{|c|}{} &
  0 &
  \multicolumn{1}{c|}{\cellcolor[HTML]{1F4E78}{\color[HTML]{F9FAFB} 0.77}} &
  \multicolumn{1}{c|}{\cellcolor[HTML]{E7ECF0}0.09} &
  \multicolumn{1}{c|}{\cellcolor[HTML]{EBEFF3}0.07} &
  \multicolumn{1}{c|}{\cellcolor[HTML]{FBFCFD}0.01} &
  \multicolumn{1}{c|}{\cellcolor[HTML]{FBFCFD}0.01} &
  \cellcolor[HTML]{F3F6F8}0.04 \\ \cline{2-8} 
\multicolumn{1}{|c|}{} &
  30 &
  \multicolumn{1}{c|}{\cellcolor[HTML]{EBEFF3}0.07} &
  \multicolumn{1}{c|}{\cellcolor[HTML]{5E809E}{\color[HTML]{F9FAFB} 0.56}} &
  \multicolumn{1}{c|}{\cellcolor[HTML]{C5D2DC}0.20} &
  \multicolumn{1}{c|}{\cellcolor[HTML]{D6DFE6}0.14} &
  \multicolumn{1}{c|}{\cellcolor[HTML]{F7F9FA}0.03} &
  \cellcolor[HTML]{F9FAFB}0.00 \\ \cline{2-8} 
\multicolumn{1}{|c|}{} &
  60 &
  \multicolumn{1}{c|}{\cellcolor[HTML]{E7ECF0}0.09} &
  \multicolumn{1}{c|}{\cellcolor[HTML]{B9C8D5}0.24} &
  \multicolumn{1}{c|}{\cellcolor[HTML]{839DB5}{\color[HTML]{F9FAFB} 0.43}} &
  \multicolumn{1}{c|}{\cellcolor[HTML]{CAD5DF}0.19} &
  \multicolumn{1}{c|}{\cellcolor[HTML]{F7F9FA}0.03} &
  \cellcolor[HTML]{F7F9FA}0.03 \\ \cline{2-8} 
\multicolumn{1}{|c|}{} &
  90 &
  \multicolumn{1}{c|}{\cellcolor[HTML]{F3F6F8}0.04} &
  \multicolumn{1}{c|}{\cellcolor[HTML]{F3F6F8}0.04} &
  \multicolumn{1}{c|}{\cellcolor[HTML]{EBEFF3}0.07} &
  \multicolumn{1}{c|}{\cellcolor[HTML]{305C82}{\color[HTML]{F9FAFB} 0.71}} &
  \multicolumn{1}{c|}{\cellcolor[HTML]{E2E9EE}0.10} &
  \cellcolor[HTML]{F7F9FA}0.03 \\ \cline{2-8} 
\multicolumn{1}{|c|}{} &
  -60 &
  \multicolumn{1}{c|}{\cellcolor[HTML]{E7ECF0}0.09} &
  \multicolumn{1}{c|}{\cellcolor[HTML]{EBEFF3}0.07} &
  \multicolumn{1}{c|}{\cellcolor[HTML]{EFF2F5}0.06} &
  \multicolumn{1}{c|}{\cellcolor[HTML]{D6DFE6}0.14} &
  \multicolumn{1}{c|}{\cellcolor[HTML]{839DB5}{\color[HTML]{F9FAFB} 0.43}} &
  \cellcolor[HTML]{C1CEDA}0.21 \\ \cline{2-8} 
\multicolumn{1}{|c|}{\multirow{-6}{*}{\textit{\rotatebox{90}{Patterns}}}} &
  -30 &
  \multicolumn{1}{c|}{\cellcolor[HTML]{EFF2F5}0.06} &
  \multicolumn{1}{c|}{\cellcolor[HTML]{EFF2F5}0.06} &
  \multicolumn{1}{c|}{\cellcolor[HTML]{FBFCFD}0.01} &
  \multicolumn{1}{c|}{\cellcolor[HTML]{DEE5EB}0.11} &
  \multicolumn{1}{c|}{\cellcolor[HTML]{C5D2DC}0.20} &
  \cellcolor[HTML]{5E809E}{\color[HTML]{F9FAFB} 0.56} \\ \hline
\end{tabular}}
\vspace{-0.35cm}
\end{table}

%% file: NasaTLX.tex
\begin{table}[h!]
\centering{
\caption{NASA TLX Rating for the System with no Haptic Feedback, Downsize Method, and with the Masked Data by CNN}
\label{CM4}
\setlength{\tabcolsep}{8pt} 
\renewcommand{\arraystretch}{0.95}
\begin{tabular}{|l|c|c|c|}
\hline
 &
  \begin{tabular}[c]{@{}c@{}}No haptic\\ feedback\end{tabular} &
  \begin{tabular}[c]{@{}c@{}}Down-size \\ method\end{tabular} &
  \begin{tabular}[c]{@{}c@{}}Tactile Patterns \\ by CNN\end{tabular} \\ \hline
Mental Demand     & 13.00 & 9.78  & 6.22 \\ \hline
Physical   Demand & 8.33  & 7.56  & 6.11 \\ \hline
Temporal   Demand & 9.33  & 8.33  & 4.89 \\ \hline
Performance       & 15.89 & 13.56 & 7.22 \\ \hline
Effort            & 13.56 & 12.44 & 8.67 \\ \hline
Frustration       & 13.22 & 10.56 & 5.00 \\ \hline
\end{tabular}}

\vspace{-0.6cm}
\end{table}